\documentclass[runningheads]{llncs}
\usepackage[T1]{fontenc}
\usepackage[ruled,vlined]{algorithm2e}
\usepackage{pgffor} 
\usepackage{adjustbox} 
\usepackage{graphicx,verbatim}
\usepackage{amsmath}
\usepackage{amssymb}
\usepackage{booktabs}
\usepackage{graphicx}
\usepackage{adjustbox}
\usepackage{hyperref}
\usepackage{multirow}
\usepackage{float}
\usepackage{marvosym}
\usepackage{caption}
\usepackage{tikz}
\usepackage{calc}
\usetikzlibrary{matrix,positioning}
\begin{document}
\title{Quantification of Uncertainty with Adversarial Models in Medical Image Segmentation}
\titlerunning{QUAM-SM}
%
\author{Hana Jebril\thanks{Equal Contribution}\inst{1,2}\orcidID{0000-0002-6593-5662} \and
Thomas Pinetz\protect\footnotemark[1]\inst{1,2}\orcidID{0000-0002-6100-2136}\Letter \and
Günter Klambauer\inst{3,4}\orcidID{0000-0003-2861-5552}\and
Hrvoje Bogunović\inst{1,2}\orcidID{0000-0002-9168-0894}}
\authorrunning{H. Jebril et al.}
%
\institute{Institute of Artificial Intelligence, Center for Medical Data Science, Medical University of~Vienna, Austria \and
Comprehensive Center for AI in Medicine, Medical University of Vienna, Austria \and
ELLIS Unit Linz, LIT AI Lab and Institute for Machine Learning, Johannes Kepler University Linz, Austria \and
Clinical Research Center for Medical AI, Johannes Kepler University Linz, Austria\\
\email{hana.jebril@meduniwien.ac.at thomas.pinetz@meduniwien.ac.at klambauer@ml.jku.at hrvoje.bogunovic@meduniwien.ac.at} 
}

\newcommand{\methodname}{Quantification of Uncertainty with Adversarial Models in Medical Image Segmentation}   
\maketitle              
\begin{abstract}
Reliable pixel-level uncertainty quantification holds the potential to transform clinical workflows by enabling high-fidelity longitudinal monitoring and distinguishing true pathological changes from artifacts. Ideally, these models provide the stability required for critical treatment planning and surgical intervention. However, standard deep learning models often suffer from miscalibration, yielding overconfident predictions that mask underlying vulnerabilities at subtle pathological boundaries. To address this, we propose QUAM-SM, a post-hoc framework using targeted adversarial search to identify "adversarially fragile" pixels. By actively seeking perturbations that expose predictive instability, our method highlights regions where decisions are most vulnerable to being flipped. Importantly, the framework disentangles epistemic uncertainty from aleatoric uncertainty. Experiments on two public datasets with multiple expert annotations demonstrate that QUAM-SM outperforms both standard and recent uncertainty estimation approaches in terms of reliability and boundary sensitivity. Code is available at \href{https://github.com/HanaJebril/quam_sm}{https://github.com/HanaJebril/quam\_sm}
\keywords{Aleatoric Uncertainty \and Adversarial Search \and Segmentation.}

\end{abstract}
\section{Introduction}

The deployment of deep learning models in clinical workflows requires not only accurate segmentation but also reliable pixel-level uncertainty quantification \cite{gawlikowski2023survey}. In applications such as longitudinal lesion monitoring, small segmentation errors may lead to incorrect clinical decisions, for example, when distinguishing true progression from pseudo-progression \cite{diaz2022recent,shi2023uncertainty}.

Despite their strong performance, standard segmentation networks often produce overconfident predictions that mask underlying failure modes, particularly at object boundaries and in subtle pathological regions. Uncertainty in medical image segmentation arises from multiple sources, including limited training data and model uncertainty (epistemic uncertainty) \cite{lambert2024trustworthy}, as well as annotation variability and inherent data ambiguity (aleatoric uncertainty) \cite{jungo2018effect,liu2022variational}. A growing body of work has explored the relationship between inter-rater variability and these uncertainty components \cite{roshanzamir2023inter}, emphasizing the clinical importance of disentangling their effects. Prior work has highlighted the clinical relevance of these components and their relationship to inter-rater variability \cite{roshanzamir2023inter}. Existing approaches to improve model reliability, such as Deep Ensembles \cite{lakshminarayanan2017simplescalablepredictiveuncertainty}, Bayesian approximations (e.g., Monte Carlo Dropout) \cite{mcdrop},
and recent self-supervised evidential methods like SURE \cite{sure}, have shown promise. However, most methods rely on stochastic sampling within limited regions of the posterior, restricting their ability to capture the full spectrum of potential prediction failures.

To address this, we propose QUAM-SM, which leverages the principle that uncertain pixels are inherently easier for an adversarial agent to perturb into a false class. By conducting a post-hoc adversarial search, our method identifies "adversarially fragile" pixels where the model's decision is easily flipped, thereby providing a more sensitive map of potential errors compared to traditional density-based uncertainty estimation. We extend the concept of Quantification of Uncertainty with Adversarial Models \cite{quam} to the segmentation domain, achieving improved coverage of the posterior by highlighting pixels that are most susceptible to being "surely false". For classification tasks, targeted attacks are defined over the discrete class labels. However, in semantic segmentation, each pixel represents an independent classification instance, leading to a much larger space of targeted attack possibilities. This pixel-wise formulation allows the adversarial search to drive predictions toward diverse alternative segmentation hypotheses. Our key contributions are as follows:
\begin{enumerate}
    \item We propose a novel post-hoc uncertainty quantification framework for medical image segmentation based on adversarial models that identifies fragile pixel-wise predictions.
    \item We distinguish between aleatoric and epistemic uncertainty, and show improved consistency with inter-rater variability on multiple public datasets.
    \item We evaluate the proposed method on multiple-observer segmentation tasks, highlighting its effectiveness in capturing uncertain pixels.
\end{enumerate}

\section{Method}

\begin{figure*}[htbp]
\centering
\includegraphics[width=\linewidth]{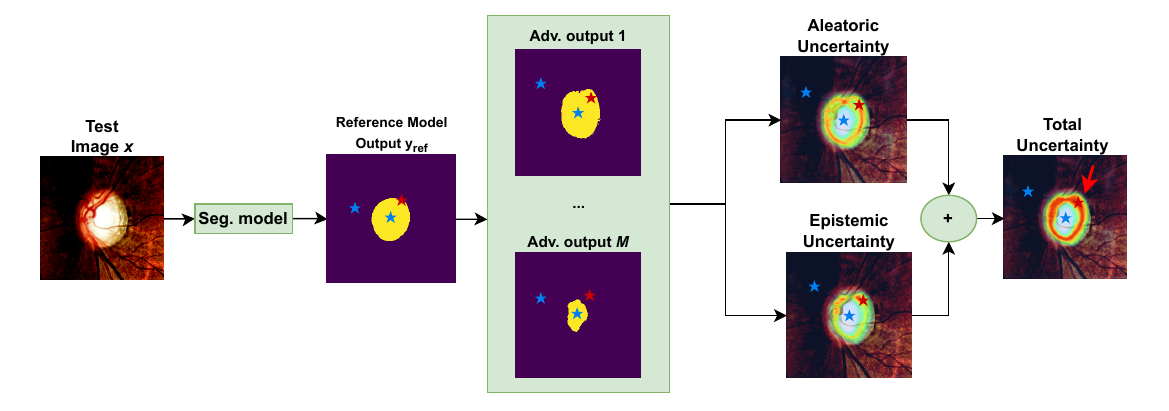}
\caption{QUAM-SM Overview: Following initial segmentation ($y_{ref}$), QUAM-SM utilizes post-hoc adversarial search across $M$ adversarial model outputs to identify predictive fragility. This fragility is decomposed into aleatoric and epistemic components and combined into a Total Uncertainty map, effectively distinguishing boundary instability (red star) from robust certain regions (blue stars).}
\label{fig:flowchart}
\end{figure*}

The \methodname\ (QUAM-SM) framework (Fig.~\ref{fig:flowchart}) first generates a reference segmentation $y_{ref}$ using a fixed pre-trained model, assuming access to the training data or training distribution to maintain consistency with the learned data manifold. We then perform a post-hoc adversarial search (Sec.~\ref{quam}) to identify "adversarially fragile" pixel regions where the model’s decision can be easily flipped to an incorrect class. Finally, uncertainty is quantified by measuring this predictive instability (Sec.~\ref{uq}), highlighting pixels most likely to be erroneous and supporting safer clinical interpretation.

\subsection{Adversarial Model Search Algorithm}\label{quam}

The adversarial search aims to generate predictions that remain consistent with the training data while diverging from the reference model on test samples, enabling broader posterior coverage and improved uncertainty characterization. To achieve this, we optimize two complementary loss terms (Alg.~\ref{alg:adv-label}). The first is a penalty loss that enforces consistency with the training distribution to ensure posterior plausibility, which requires access to training data. The second is an adversarial loss that drives the prediction toward a targeted label $y_{mask}$, encouraging exploration of alternative segmentation hypotheses.
\subsubsection{Targeted Attacks:} \label{attacks}
To alter the adversarial model predictions, adversarial attacks are applied. In a targeted attack, the optimization explicitly drives the model toward a predefined output class by minimizing the loss with respect to the desired target \cite{yu2019generatingadversarialexamplesconditional,ijcai2021p93}. In this work, we investigate multiple targeted attack strategies to broaden posterior exploration. Formally, let $y_{\text{ref}}$ be the reference prediction and $\mathcal{D}_{\text{train}}$ the set of training masks. We define three targeted adversarial labels $y_{\text{mask}}$:
\begin{enumerate}
    \item \textbf{Global extreme masks (GEM)}: representing pixel-wise constant segmentation maps where all pixels belong to either the background or the foreground, which is a straight-forward extension to QUAM approach \cite{quam}: $y_{\text{mask}} \in \left\{ \mathbf{0}_{H \times W},\;\mathbf{1}_{H \times W} \right\}$

    \item \textbf{Training-based masks (TBM)}: encouraging the adversarial model to explore realistic anatomical configurations; $y_{\text{mask}} \sim \mathcal{D}_{\text{train}} \quad \text{(\textit{$\omega$} selected masks)}$.

    \item \textbf{Morphology-augmented masks (MAM)}: $\mathcal{M}(\cdot)$ denotes a morphological operation such as dilation or erosion, probing local structural variability near boundaries which are eliminated; $y_{\text{mask}} = \mathcal{M}(y_{\text{ref}})$

\end{enumerate}

\begin{algorithm}[htbp]
\small
\caption{Adversarial Model Search Algorithm}
\label{alg:adv-label}
\KwIn{Test image $x$, training set $D$, reference model $w$, reference loss $\mathcal{L}_{ref}$,targeted label $y_{mask}$, penalty scheduler $\eta$, divergence $D(\cdot,\cdot)$, parameters $c_0, M, \gamma$, Compute $y_{ref} = p(y|x, w)$}
\KwOut{Optimized adversarial model $\tilde{w}$}

Initialize $\tilde{w} \gets w$, $c \gets c_0$ \;

\For{$m = 1$ \KwTo $M$}{
    Compute penalty loss: 
    $\displaystyle \mathcal{L}_{pen} = \frac{1}{K} \sum_{i=1}^{K} \big[ \ell(p(y_i|x_i,\tilde{w}), y_i) - (\mathcal{L}_{ref} + \gamma) \big]$ \;
    Compute $y_{\text{mask}} = \mathcal{M}(y_{\text{ref}})$\;
    Compute adversarial loss:
    $\displaystyle \mathcal{L}_{adv} = - D(y_{mask}, p(y|x, \tilde{w}))$ \;
    
    Total loss:
    $\displaystyle \mathcal{L} = \mathcal{L}_{adv} + c \,\mathcal{L}_{pen}$ \;
    
    Update parameters: $\tilde{w} \gets \eta(\tilde{w}), \; c \gets \eta(c)$ \;
    
    \If{$\mathcal{L}_{pen} \le 0$ and $\mathcal{L}_{adv}$ improved}{store $\tilde{w}$ and break \;}
}\Return $\tilde{w}$ \;
\end{algorithm}

\subsection{Predictive Instability Uncertainty Quantification}\label{uq}

The output of the adversarial search, a set of plausible model weights $\{\mathbf{w}_t\}_{t=1}^N$, forms a sparse pseudo-posterior from which we estimate the predictive distribution $P(Y \mid \mathbf{x}, \mathcal{D})$.

\subsubsection{Mixture Importance Sampling (MIS)}
Due to the adversarial bias in sample generation, we employ Mixture Importance Sampling (MIS) \cite{hesterberg1995weighted} to derive a statistically consistent estimate of the expectation $\mathbb{E}_{\mathbf{w}}[\cdot]$ over the true posterior $P(\mathbf{w} \mid \mathcal{D})$. The expectation is approximated by a weighted sum over the sampled models $\mathbf{w}_t$:
\begin{equation}
\mathbb{E}_{\mathbf{w}}[\mathcal{F}(\mathbf{w})] \approx \sum_{t=1}^{N} W_t \cdot \mathcal{F}(\mathbf{w}_t)
\quad \text{where} \quad
W_t = \frac{\exp(-\mathcal{L}_{\text{pen}}(\mathbf{w}_t) / T)}{\sum_{j=1}^{N} \exp(-\mathcal{L}_{\text{pen}}(\mathbf{w}_j) / T)}
\end{equation}
The weight $W_t$ is the MIS importance weight derived from the model's Penalty Loss ($\mathcal{L}_{\text{pen}}$), following the original QUAM work~\cite{quam}. In particular, the term $\exp(-\mathcal{L}_{\text{pen}}(\mathbf{w}_t) / T)$ ensures that models $\mathbf{w}_t$ which achieve low penalized loss on the training data contribute proportionally more to the final predictive distribution. $T$ is the temperature hyperparameter controlling the sharpness of the weight distribution. We define the Bayesian Model Average (BMA) weighted prediction $P_{\mathcal{D}} = \sum_{t} W_t P(Y \mid \mathbf{x}, \mathbf{w}_t)$. The resulting Weighted BMA prediction yields more robust and accurate estimates for both Aleatoric and Epistemic uncertainty.

\subsubsection{Total Uncertainty Decomposition}
We decompose the total predictive uncertainty $H[P_{\mathcal{D}}]$ into its aleatoric and epistemic components on a per-pixel basis using the established Expected Entropy (EE) framework. This standard decomposition leverages the properties of Mutual Information and conditional entropy, where the Epistemic Uncertainty is identified with the Mutual Information between the predicted outcome $Y$ and the model parameters $\mathbf{w}$:

\begin{equation} \label{eq:ee_decomposition_weighted}
H[P_{\mathcal{D}}] = \underbrace{\mathbb{E}_{\mathbf{w}}\left[H[P(Y \mid \mathbf{x}, \mathbf{w})]\right]}_{\text{Aleatoric Uncertainty}} + \underbrace{\mathbb{I}[Y ; \mathbf{w} \mid \mathbf{x}, \mathcal{D}]}_{\text{Epistemic Uncertainty}}
\end{equation}

Let $N$ be the number of sampled model weights, $\hat{P}_{t} \in \mathbb{R}^C$ the softmax probability vector from the $t$-th model, and $W_t$ its corresponding MIS weight ($\sum_{t=1}^{N} W_t = 1$). $C$ is the total number of segmented classes. The uncertainty components are estimated using the weighted expectations:

\begin{itemize}
    \item \textbf{Total Uncertainty ($\mathbf{H}_{\text{Total}}$)}: The entropy of the Weighted Bayesian Model Averaging (BMA) prediction.
    \begin{equation} \label{eq:total_unc_weighted}
    \mathbf{H}_{\text{Total}} = - \sum_{c=1}^{C} \left( \sum_{t=1}^{N} W_t \cdot \hat{P}_{t,c} \right) \log \left( \sum_{t=1}^{N} W_t \cdot \hat{P}_{t,c} \right)
    \end{equation}

    \item \textbf{Aleatoric Uncertainty ($\mathbf{H}_{\text{Aleatoric}}$)}: The expected entropy of the individual predictions, estimated by the weighted average of individual entropies.
    \begin{equation} \label{eq:aleatoric_unc_weighted}
    \mathbf{H}_{\text{Aleatoric}} = \sum_{t=1}^{N} W_t \cdot \left( - \sum_{c=1}^{C} \hat{P}_{t,c} \log(\hat{P}_{t,c}) \right)
    \end{equation}

    \item \textbf{Epistemic Uncertainty ($\mathbf{H}_{\text{Epistemic}}$)}: Calculated as the weighted measure of disagreement (KL divergence).
    \begin{equation} \label{eq:epistemic_unc_weighted}
    \mathbf{H}_{\text{Epistemic}} = \sum_{t=1}^{N} W_t \cdot \left[ \sum_{c=1}^{C} \hat{P}_{t,c} \log \left( \frac{\hat{P}_{t,c}}{\sum_{k=1}^{N} W_k \cdot \hat{P}_{k,c}} \right) \right] \end{equation}
\end{itemize}

\section{Experiments and Results}\label{er}

\subsection{Datasets and Implementation Details}\label{dset}
To evaluate our proposed QUAM-SM, we use two public datasets with multiple annotations available. 
(1) The REFUGE dataset ($\mathcal{D}_R$) \cite{Orlando_2020} consists of 400 color fundus photography (CFP) images for training, 400 for validation, and 400 for testing. Each image is accompanied by seven independent segmentation annotations of the optic disc and optic cup, provided by multiple expert observers. 
(2) The QUBIQ2021 prostate tumour dataset ($\mathcal{D}_Q$) \cite{becker2019variability} consists of 45 magnetic resonance imaging (MRI) volumes for training and 7 volumes for testing, each has one selected 2D slice and annotated by six experts. 
All images were cropped to a spatial resolution of $512 \times 512$ before training and evaluation.

\subsubsection{Comparison Study}
We compare the proposed QUAM-SM against several widely used uncertainty-aware segmentation baselines, namely Monte Carlo (MC) Dropout~\cite{mcdrop}, Deep Ensembles (DE)~\cite{lakshminarayanan2017simplescalablepredictiveuncertainty}, Probabilistic U-Net (PUnet)~\cite{kohl2018probabilistic}, and the Uncertainty-Supervised Interpretable and Robust Evidential Segmentation (SURE) framework~\cite{sure}. For MC, DE, and PUnet, aleatoric and epistemic uncertainty maps are computed using the formulations in Eqs.~(\ref{eq:total_unc_weighted}–\ref{eq:epistemic_unc_weighted}), while omitting the proposed weighting term.

For the evidential-based SURE model, uncertainty estimation follows the formulation introduced in~\cite{tan2024uncertainty}. Specifically, let $\boldsymbol{\alpha} = \{\alpha_i\}_{i=1}^{K}$ denote the Dirichlet concentration parameters over $K$ classes and $S=\sum_{i=1}^{K}\alpha_i$. Aleatoric and epistemic uncertainties are computed as:

\begin{equation}
\begin{aligned}
\mathbf{u}_{\text{aleatoric}} &= \sum_{i=1}^{K} \frac{\alpha_i (S - \alpha_i)}{S(S+1)}, \quad
\mathbf{u}_{\text{epistemic}} &= \sum_{i=1}^{K} \frac{\alpha_i (S - \alpha_i)}{S^2(S+1)}.
\end{aligned}
\end{equation}

\subsubsection{Implementation details:} We employed the Attention U-Net \cite{oktay2018attentionunetlearninglook} as a backbone for our algorithm and other baselines, and Adam optimizer with a learning rate of $10^{-4}$. The batch size was set to $4$ for both datasets.
To evaluate the quality of uncertainty estimation, we measure the agreement between the predicted uncertainty maps and the reference entropy map derived from multiple annotators. We report the Pearson Correlation Coefficient (PCC) and the coefficient of determination ($R^2$) to assess correlation and goodness of fit. In addition, the Soft Dice (SDice) is used to evaluate the segmentation.

\subsection{Experimental Results}\label{merged}
\begin{figure}[htbp]
\centering
\noindent 
\resizebox{\textwidth}{!}{%
\begin{tikzpicture}[
    node distance = 0.1cm,
    image style/.style={inner sep=0pt, anchor=north west},
    label style/.style={align=center, anchor=north}
]

    \def\imgwidth{1.6cm}  
    \def\hgap{0.1}        
    \def\vgap{0.1}        
    \def\groupgap{0.5}    
    \def\basepath{figures}

    \foreach \unx [count=\i] in {ale, tot} {
        \foreach \idx [count=\j] in {50,0484} {            
            \foreach \mdx [count=\c from 0] in {input, label, de/\unx, mc/\unx, punet/\unx, sure/\unx, quam/\unx} {
                
                \pgfmathsetmacro{\xpos}{\c * (\imgwidth/1cm + \hgap)}
                \pgfmathsetmacro{\ypos}{2 * \i * (\imgwidth / 1cm + 3*\vgap) + \j *  (\imgwidth / 1cm + \vgap) }
                \node[image style] (img-\i-\unx-\c) at (\xpos, \ypos) {
                    \includegraphics[width=\imgwidth]{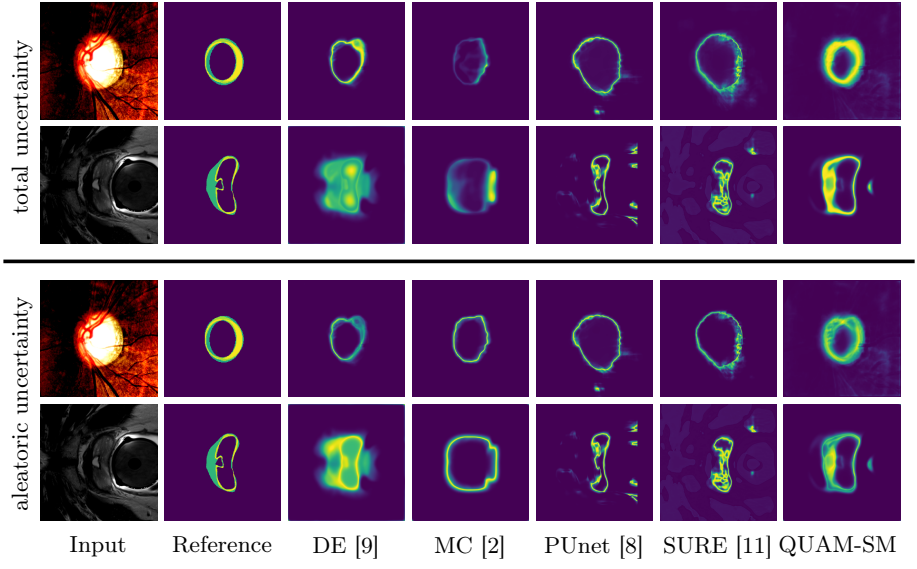}
                };
            }
        }
    }

    \node[label style, rotate=90] at (-0.5, 9.3) {total uncertainty};
    \node[label style, rotate=90] at (-0.5, 5.5) {aleatoric uncertainty};
    \draw[line width=0.5 mm] (-0.5,7.45) -- (12,7.45);

    \foreach \lab [count=\c from 0] in {
        Input, 
        Reference, 
        DE~\cite{lakshminarayanan2017simplescalablepredictiveuncertainty}, 
        MC~\cite{mcdrop}, 
        PUnet~\cite{kohl2018probabilistic}, 
        SURE~\cite{sure}, 
        QUAM-SM
    } {
        \pgfmathsetmacro{\xpos}{\c * (\imgwidth/1cm + \hgap) + (\imgwidth/1cm)/2}
        \node[label style] at (\xpos, 3.8) {\lab};
    }

\end{tikzpicture}
}
\caption{Qualitative comparison between our method and state-of-the-art for total and aleatoric uncertainty on the same cases.}
\label{fig:merged_vis}
\end{figure}

Table~\ref{table:merged_uncertainty_comparison} summarizes the quantitative results on the multi-annotator datasets REFUGE ($\mathcal{D}_R$) and QUBIQ2021 prostate tumour ($\mathcal{D}_Q$). We compare different uncertainty estimation methods by computing the correlation between each predicted uncertainty map and the multi-observer entropy map. Our method (QUAM-SM) achieves the highest PCC and $R^2$ values across epistemic, aleatoric, and total uncertainty, indicating superior alignment with the reference uncertainty. In terms of segmentation accuracy, QUAM-SM method also consistently outperforms all competing approaches for all uncertainty types, as reflected by the SDice scores. Notably, the improvement for both datasets is most pronounced for aleatoric uncertainty, which is expected since the multiple annotation task primarily reflects aleatoric uncertainty arising from inter-observer variability. Figure~\ref{fig:merged_vis} presents representative qualitative results from each dataset for both aleatoric and total uncertainty. The visualizations demonstrate that QUAM-SM closely matches the ground-truth entropy maps, accurately capturing regions of high inter-annotator uncertainty.

\begin{table*}[htbp]
\centering
\caption{Comparison of uncertainty methods $\mathcal{M}$ (DE~\cite{lakshminarayanan2017simplescalablepredictiveuncertainty}, MC~\cite{mcdrop}, PUnet~\cite{kohl2018probabilistic}, SURE~\cite{sure}, TTA~\cite{tta2}) on REFUGE ($\mathcal{D}_R$) and QUBIQ2021 ($\mathcal{D}_Q$).}
\label{table:merged_uncertainty_comparison}
\resizebox{\textwidth}{!}{
\begin{tabular}{l|l|
ccc|
ccc|
ccc}
\toprule
\multirow{2}{*}{$\mathcal{D}$} & \multirow{2}{*}{$\mathcal{M}$}
& \multicolumn{3}{c|}{Epistemic}
& \multicolumn{3}{c|}{Aleatoric}
& \multicolumn{3}{c}{Total} \\
\cmidrule(r){3-5} \cmidrule(r){6-8} \cmidrule(l){9-11}
&
& R$^2$ & PCC & SDice
& R$^2$ & PCC & SDice
& R$^2$ & PCC & SDice \\
\midrule
\multirow{5}{*}{$\mathcal{D}_R$}
& ~\cite{lakshminarayanan2017simplescalablepredictiveuncertainty} & $.18\pm.07$ & $.41\pm.10$ & $.23\pm.07$
& $.39\pm.10$ & $.62\pm.09$ & $.55\pm.08$
& $.34\pm.10$& $.57\pm.11$ & $.43\pm.08$ \\

& ~\cite{mcdrop} & $.28\pm.11$ & $.52\pm .12$ & $.33\pm.09$
& $.39\pm.10$ & $.62\pm.10$ & $.51\pm.09$
& $.36\pm.11$ & $.59\pm.11$ & $.42\pm.09$ \\

& ~\cite{kohl2018probabilistic}& $.10\pm.11$ & $.25\pm.20$ & $.30\pm.15$
& $.13\pm.11$ & $.31\pm.17$ & $.30\pm.15$
& $.13\pm.11$ & $.31\pm.17$ & $.29\pm.14$ \\

& ~\cite{sure} & $.07 \pm .11$ & $.20 \pm .18$ & $.25\pm.1$
& $.08 \pm .09$ & $.18 \pm .06$ & $.23 \pm .11$
& $.08\pm .10$ & $.19 \pm .16$ & $.24\pm.09$ \\

& ~\cite{tta2} & $.18 \pm .07$ & $.41 \pm .11$ & $.19\pm.06$
& $.40 \pm .11$ & $.62 \pm .10$ & $.52 \pm .08$
& $.36\pm .11$ & $.59 \pm .11$ & $.42\pm.08$ \\

& Ours & $\mathbf{.54\pm.11}$ & $\mathbf{.73 \pm .09}$ & $\mathbf{.73\pm.08}$
& $\mathbf{.64 \pm .11}$ & $\mathbf{.80\pm.08}$ & $\mathbf{.80\pm.07}$
& $\mathbf{.63\pm.11}$ & $\mathbf{.79\pm.07}$ & $\mathbf{.78\pm.08}$ \\
\midrule
\multirow{5}{*}{$\mathcal{D}_Q$}
& ~\cite{lakshminarayanan2017simplescalablepredictiveuncertainty} & $.04\pm.02$ & $.17\pm.08$ & $.20\pm.07$
& $.21\pm .09$ & $.44\pm.12$ & $.41\pm.11$
& $.17\pm .08$ & $.39\pm.12$ & $.33\pm.10$ \\
& ~\cite{mcdrop} & $.10\pm.09$ & $.29\pm.14$ & $.32\pm.11$
& $.12\pm.07$ & $.33 \pm .10$ & $.28\pm.09$
& $.12\pm.09$ &  $.34\pm.11$ & $.33\pm.11$ \\
& ~\cite{kohl2018probabilistic} & $.08\pm.07$ & $.27\pm.12$ & $.15\pm.07$
& $.14\pm.08$ & $.36\pm.10$ & $.38\pm.09$
& $.14\pm.08$ & $.36\pm.10$ & $.38\pm.09$ \\
& ~\cite{sure} & $.13\pm.15$ & $.31\pm.20$ & $.35\pm.17$
& $.13\pm.14$ & $.31\pm.20$ & $.34\pm.18$
& $.13\pm.14$ & $.31\pm.20$ & $.35\pm.18$ \\

& ~\cite{tta2} & $.15 \pm .11$ & $.37 \pm .13$ & $.38\pm.10$
& $.37 \pm .12$ & $.60 \pm .11$ & $.59 \pm .09$
& $.33\pm .10$ & $.57 \pm .10$ & $.54\pm.09$ \\

& Ours & $\mathbf{.20\pm.09}$ & $\mathbf{.44\pm.13}$ & $\mathbf{.39\pm.13}$
& $\mathbf{.40\pm.16}$ & $\mathbf{.62\pm.14}$ & $\mathbf{.62\pm.13}$
& $\mathbf{.36\pm.13}$ & $\mathbf{.58\pm.13}$ & $\mathbf{.57\pm.13}$ \\
\bottomrule
\end{tabular}
}
\end{table*}

\subsection{Ablation Study}
We evaluate the effectiveness of three targeted adversarial attacks introduced in Sec.~\ref{attacks}. As shown in Table~\ref{table:kernel}, top part, the targeted attack based on morphological operations consistently outperforms the other attack strategies across epistemic, aleatoric, and total uncertainty measures.
We also investigate the impact of the structuring element (kernel) size and the number of iterations in the morphology-based targeted attack. Table~\ref{table:kernel}, second part, shows that kernel sizes of $3$ and $5$ yield the best performance across uncertainty maps, with erosion and dilation applied for three iterations. Furthermore, increasing the number of iterations $M$ in the QUAM-SM algorithm (Sec.~\ref{quam}) consistently improves performance as shown in the last part of Table~\ref{table:kernel}. The maximum number of iterations corresponds to the length of the training loader, i.e., $M = (\text{len(train-loader)}/x) \times D$, where $x \in \{1, \ldots, \lvert \text{train-loader} \rvert\}$ and $D$ denotes the number of adversarial models. 
For the REFUGE ($\mathcal{D}_R$) dataset, we set $\lvert \text{train-loader} \rvert = 100$ and $D = 2$.

\begin{table}[htbp]
\centering
 
\caption{Ablation study for the selection of \textit{kernel size} $k$ and $M$ on REFUGE ($\mathcal{D}_R$). We set $\omega=5$ for \textbf{TBM}.}
\label{table:kernel}
\resizebox{\textwidth}{!}{
\begin{tabular}{c|
ccc|
ccc|
ccc}

\toprule
$\mathbf{y}_{\mathbf{mask}}$
& \multicolumn{3}{c|}{Epistemic}
& \multicolumn{3}{c|}{Aleatoric}
& \multicolumn{3}{c}{Total} \\
\cmidrule(r){2-4} \cmidrule(r){5-7} \cmidrule(l){8-10}

& R$^2$  & PCC  & SDice 
& R$^2$  & PCC  & SDice 
& R$^2$  & PCC  & SDice  \\
\midrule
\textbf{GEM}   
& $.13\pm.09$ & $.33\pm.14$ & $.29\pm.12$
& $.33\pm.15$ & $.56\pm.14$ & $.56\pm.12$
& $.33\pm.15$ & $.56\pm.14$ & $.56\pm.12$ \\
\textbf{TBM}
& $.12\pm.10$ & $.32\pm.14$ & $.27\pm.11$
& $.34\pm.16$ & $.56\pm.14$ & $.57\pm.13$
& $.32\pm.16$ & $.55\pm.14$ & $.56\pm.13$ \\
\textbf{MAM}
& $\mathbf{.38\pm.13}$ & $\mathbf{.61\pm.11}$  & $\mathbf{.62\pm.10}$ 
& $\mathbf{.51\pm.13}$ & $\mathbf{.71\pm.09}$ & $\mathbf{.71\pm.08}$ 
& $\mathbf{.50\pm.13}$ & $\mathbf{.70\pm.09}$ & $\mathbf{.69\pm.10}$ \\

\toprule
$k$
& \multicolumn{3}{c|}{Epistemic}
& \multicolumn{3}{c|}{Aleatoric}
& \multicolumn{3}{c}{Total} \\
\midrule

1
& $.33\pm.14$ & $.56\pm.13$ & $.53\pm.11$
& $.52\pm.13$ & $.71\pm.10$ & $.70\pm.08$
& $.49\pm.14$ & $.69\pm.11$ & $.70\pm.11$ \\

3
& $\mathbf{.39\pm.14}$ & $\mathbf{.61\pm.12}$ & $.61\pm.10$
& $\mathbf{.53\pm.13}$ & $\mathbf{.72\pm.09}$ & $\mathbf{.72\pm.08}$
& $\mathbf{.52\pm.14}$ & $\mathbf{.71\pm.10}$ & $\mathbf{.71\pm.11}$ \\

5
& $.38\pm.13$ & $.60\pm.11$ & $\mathbf{.62\pm.10}$ 
& $.51\pm.13$ & $.71\pm.09$ & $.71\pm.08$ 
& $.50\pm.13$ &$.70\pm.09$ & $.69\pm.10$ \\

7 
& $.35\pm.12$ & $.59\pm.10$ & $.60\pm.09$
& $.47\pm.12$ & $.68\pm.09$ & $.69\pm.09$
& $.47\pm.11$ & $.68\pm.08$ & $.66\pm.10$ \\

9
& $.33\pm.11$ & $.56\pm.09$ & $.58\pm.09$
& $.44\pm.13$ & $.65\pm.10$ & $.66\pm.09$
& $.44\pm.10$ & $.66\pm.08$ & $.63\pm.10$ \\

\toprule
\textbf{\textit{M}}
& \multicolumn{3}{c|}{Epistemic}
& \multicolumn{3}{c|}{Aleatoric}
& \multicolumn{3}{c}{Total} \\
\midrule

2 
& $.33\pm.13$ & $.56\pm.12$ & $.56\pm.11$
& $.47\pm.14$ & $.67\pm.11$ & $.66\pm.09$
& $.46\pm.13$ & $.67\pm.10$ & $.66\pm.10$ \\

4 
& $.39\pm.13$ & $.61\pm.11$ & $.61\pm.10$
& $.53\pm.13$ & $.72\pm.09$ & $.71\pm.08$
& $.51\pm.13$ & $.71\pm.09$ & $.70\pm.10$ \\

10 
& $.45\pm.13$ & $.67\pm.09$ & $.66\pm.08$ 
& $.60\pm.11$ & $.77\pm.07$ & $.77\pm.07$ 
& $.58\pm.11$ &$.76\pm.07$ & $.76\pm.08$ \\

20 
& $.48\pm.13$ & $.69\pm.10$ & $.69\pm.09$
& $.60\pm.11$ & $.77\pm.07$ & $.78\pm.07$
& $.58\pm.11$ & $.76\pm.07$ & $.75\pm.08$ \\

40 
& $.53\pm.12$ & $.72\pm.09$ & $.72\pm.08$
& $\mathbf{.64\pm.10}$ & $\mathbf{.80\pm.07}$ & $\mathbf{.80\pm.07}$
& $\mathbf{.63\pm.11}$ & $\mathbf{.79\pm.07}$ & $\mathbf{.78\pm.07}$ \\

100 
& $.53\pm.12$ & $.72\pm.08$ & $.72\pm.08$
& $.63\pm.10$ & $.79\pm.07$ & $.79\pm.07$
& $.62\pm.10$ & $.78\pm.07$ & $.77\pm.07$ \\

200 
& $\mathbf{.54\pm.12}$ & $\mathbf{.73\pm.08}$ & $\mathbf{.73\pm.08}$
& $.63\pm.11$ & $.79\pm.07$ & $.79\pm.07$
& $.61\pm.10$ & $.78\pm.07$ & $.77\pm.07$ \\
\bottomrule
\end{tabular}
}
\end{table}











\section{Conclusions}
In this paper, we presented QUAM-SM, a post-hoc uncertainty quantification framework for medical image segmentation based on targeted adversarial search. The method identifies adversarially fragile pixels to reveal prediction instability and mitigate the clinical risks of overconfident model decisions. QUAM-SM produces interpretable epistemic and aleatoric uncertainty maps for more reliable model assessment. While the approach requires access to the training distribution, experiments on realistic clinical benchmarks show that the estimated aleatoric uncertainty correlates more strongly with inter-rater variability than total uncertainty, highlighting its relevance for capturing annotation ambiguity and supporting safer clinical decision-making.

\subsubsection{Acknowledgments} This work was supported in part by the Austrian Science Fund (FWF) grant 10.55776/FG9.
TP is funded by the European Union (I-SCREEN, grant no. 101130093), EIC-2023-PATHFINDEROPEN-01. Views and opinions expressed are however those of the author(s) only and do not necessarily reflect those of the European Union or European Innovation Council and SMEs Executive Agency (EISMEA). Neither the European Union nor the granting authority can be held responsible for them.

\subsubsection{Disclosure of Interests}  The authors have no competing interests to declare.

%
%
%
\bibliographystyle{splncs04}
\bibliography{references}

@article{gawlikowski2023survey,
  title={A survey of uncertainty in deep neural networks},
  author={Gawlikowski, Jakob and Tassi, Cedrique Rovile Njieutcheu and Ali, Mohsin and Lee, Jongseok and Humt, Matthias and Feng, Jianxiang and Kruspe, Anna and Triebel, Rudolph and Jung, Peter and Roscher, Ribana and others},
  journal={Artificial Intelligence Review},
  volume={56},
  number={Suppl 1},
  pages={1513--1589},
  year={2023},
  publisher={Springer}
}

@article{lambert2024trustworthy,
  title={Trustworthy clinical {AI} solutions: A unified review of uncertainty quantification in Deep Learning models for medical image analysis.},
  author={Lambert, Benjamin and Forbes, Florence and Doyle, Senan and Dehaene, Harmonie and Dojat, Michel},
  journal={Artif. Intell. Medicine},
  volume={150},
  pages={102830},
  year={2024},
  publisher={Elsevier}
}

@article{shi2023uncertainty,
  title={Uncertainty-weighted and relation-driven consistency training for semi-supervised head-and-neck tumor segmentation},
  author={Shi, Yuang and Zu, Chen and Yang, Pinli and Tan, Shuai and Ren, Hongping and Wu, Xi and Zhou, Jiliu and Wang, Yan},
  journal={Knowledge-Based Systems},
  volume={272},
  pages={110598},
  year={2023},
  publisher={Elsevier}
}

@inproceedings{jungo2018effect,
  title={On the effect of inter-observer variability for a reliable estimation of uncertainty of medical image segmentation},
  author={Jungo, Alain and Meier, Raphael and Ermis, Ekin and Blatti-Moreno, Marcela and Herrmann, Evelyn and Wiest, Roland and Reyes, Mauricio},
  booktitle={International Conference on Medical Image Computing and Computer-Assisted Intervention (MICCAI)},
  pages={682--690},
  year={2018},
  organization={Springer}
}

@inproceedings{liu2022variational,
  title={Variational inference for quantifying inter-observer variability in segmentation of anatomical structures},
  author={Liu, Xiaofeng and Xing, Fangxu and Marin, Thibault and El Fakhri, Georges and Woo, Jonghye},
  booktitle={Medical Imaging 2022: Image Processing},
  volume={12032},
  pages={438--443},
  year={2022},
  organization={SPIE}
}

@article{lakshminarayanan2017simplescalablepredictiveuncertainty,
  title={Simple and scalable predictive uncertainty estimation using deep ensembles},
  author={Lakshminarayanan, Balaji and Pritzel, Alexander and Blundell, Charles},
  journal={Advances in neural information processing systems},
  volume={30},
  year={2017}
}

@InProceedings{mcdrop,
author="Camarasa, Robin
and Bos, Daniel
and Hendrikse, Jeroen
and Nederkoorn, Paul
and Kooi, Eline
and van der Lugt, Aad
and de Bruijne, Marleen",
editor="Sudre, Carole H.
and Fehri, Hamid
and Arbel, Tal
and Baumgartner, Christian F.
and Dalca, Adrian
and Tanno, Ryutaro
and Van Leemput, Koen
and Wells, William M.
and Sotiras, Aristeidis
and Papiez, Bartlomiej
and Ferrante, Enzo
and Parisot, Sarah",
title="Quantitative Comparison of Monte-Carlo Dropout Uncertainty Measures for Multi-class Segmentation",
booktitle="Uncertainty for Safe Utilization of Machine Learning in Medical Imaging, and Graphs in Biomedical Image Analysis",
year="2020",
publisher="Springer International Publishing",
address="Cham",
pages="32--41",
isbn="978-3-030-60365-6"
}

@InProceedings{sure,
author="Li, Yuzhu
and Sui, An
and Wu, Fuping
and Zhuang, Xiahai",
title="Uncertainty-Supervised Interpretable and Robust Evidential Segmentation",
booktitle="Medical Image Computing and Computer Assisted Intervention (MICCAI)",
year="2026",
publisher="Springer Nature Switzerland",
address="Cham",
pages="649--658",
isbn="978-3-032-05185-1"
}

@article{quam,
  title={Quantification of uncertainty with adversarial models},
  author={Schweighofer, Kajetan and Aichberger, Lukas and Ielanskyi, Mykyta and Klambauer, G{\"u}nter and Hochreiter, Sepp},
  journal={Advances in Neural Information Processing Systems},
  volume={36},
  pages={19446--19484},
  year={2023}
}

@INPROCEEDINGS{yu2019generatingadversarialexamplesconditional,
  author={Yu, Ping and Song, Kaitao and Lu, Jianfeng},
  booktitle={24th International Conference on Pattern Recognition (ICPR)}, 
  title={Generating Adversarial Examples With Conditional Generative Adversarial Net}, 
  year={2018},
  pages={676-681},
  doi={10.1109/ICPR.2018.8545152}}

@inproceedings{ijcai2021p93,
  title     = {Feature Space Targeted Attacks by Statistic Alignment},
  author    = {Gao, Lianli and Cheng, Yaya and Zhang, Qilong and Xu, Xing and Song, Jingkuan},
  booktitle = {Proceedings of the Thirtieth International Joint Conference on Artificial Intelligence (IJCAI)},
  editor    = {Zhi-Hua Zhou},
  pages     = {671--677},
  year      = {2021},
  month     = {8},
  url       = {https://doi.org/10.24963/ijcai.2021/93},
}

@inproceedings{tan2024uncertainty,
  title={Uncertainty-Error correlations in Evidential Deep Learning models for biomedical segmentation},
  author={Tan, Hai Siong and Wang, Kwancheng and Mcbeth, Rafe},
  booktitle={International Conference on Technologies and Applications of Artificial Intelligence},
  pages={91--105},
  year={2024},
  organization={Springer}
}

@article{kohl2018probabilistic,
  title={A probabilistic {U-Net} for segmentation of ambiguous images},
  author={Kohl, Simon and Romera-Paredes, Bernardino and Meyer, Clemens and De Fauw, Jeffrey and Ledsam, Joseph R and Maier-Hein, Klaus and Eslami, SM and Jimenez Rezende, Danilo and Ronneberger, Olaf},
  journal={Advances in neural information processing systems},
  volume={31},
  year={2018}
}

@article{hesterberg1995weighted,
  title={Weighted average importance sampling and defensive mixture distributions},
  author={Hesterberg, Tim},
  journal={Technometrics},
  volume={37},
  number={2},
  pages={185--194},
  year={1995},
  publisher={Taylor \& Francis}
}

@article{Orlando_2020,
   title={{REFUGE Challenge: A unified framework for evaluating automated methods for glaucoma assessment from fundus photographs}},
   volume={59},
   ISSN={1361-8415},
   url={http://dx.doi.org/10.1016/j.media.2019.101570},
   journal={Medical Image Analysis},
   publisher={Elsevier BV},
   author={Orlando, José Ignacio and Fu, Huazhu and Barbosa Breda, João and van Keer, Karel and Bathula, Deepti R. and Diaz-Pinto, Andrés and Fang, Ruogu and Heng, Pheng-Ann and Kim, Jeyoung and Lee, JoonHo and Lee, Joonseok and Li, Xiaoxiao and Liu, Peng and Lu, Shuai and Murugesan, Balamurali and Naranjo, Valery and Phaye, Sai Samarth R. and Shankaranarayana, Sharath M. and Sikka, Apoorva and Son, Jaemin and van den Hengel, Anton and Wang, Shujun and Wu, Junyan and Wu, Zifeng and Xu, Guanghui and Xu, Yongli and Yin, Pengshuai and Li, Fei and Zhang, Xiulan and Xu, Yanwu and Bogunović, Hrvoje},
   year={2020},
   month=jan, pages={101570} }

@article{becker2019variability,
  title={{Variability of manual segmentation of the prostate in axial T2-weighted MRI: A multi-reader study}},
  author={Becker, Anton S and Chaitanya, Krishna and Schawkat, Khoschy and Muehlematter, Urs J and H{\"o}tker, Andreas M and Konukoglu, Ender and Donati, Olivio F},
  journal={European Journal of Radiology},
  volume={121},
  pages={108716},
  year={2019},
  publisher={Elsevier}
}

@inproceedings{
oktay2018attentionunetlearninglook,
title={Attention {U-Net}: Learning Where to Look for the Pancreas},
author={Ozan Oktay and Jo Schlemper and Loic Le Folgoc and Matthew Lee and Mattias Heinrich and Kazunari Misawa and Kensaku Mori and Steven McDonagh and Nils Y Hammerla and Bernhard Kainz and Ben Glocker and Daniel Rueckert},
booktitle={Medical Imaging with Deep Learning},
year={2018},
url={https://openreview.net/forum?id=Skft7cijM}
}

@article{diaz2022recent,
  title={Recent advances in the longitudinal segmentation of multiple sclerosis lesions on magnetic resonance imaging: a review},
  author={Diaz-Hurtado, Marcos and Mart{\'\i}nez-Heras, Eloy and Solana, Elisabeth and Casas-Roma, Jordi and Llufriu, Sara and Kanber, Baris and Prados, Ferran},
  journal={Neuroradiology},
  volume={64},
  number={11},
  pages={2103--2117},
  year={2022},
  publisher={Springer}
}

@inproceedings{roshanzamir2023inter,
  title={How inter-rater variability relates to aleatoric and epistemic uncertainty: a case study with deep learning-based paraspinal muscle segmentation},
  author={Roshanzamir, Parinaz and Rivaz, Hassan and Ahn, Joshua and Mirza, Hamza and Naghdi, Neda and Anstruther, Meagan and Batti{\'e}, Michele C and Fortin, Maryse and Xiao, Yiming},
  booktitle={International workshop on uncertainty for safe utilization of machine learning in medical imaging (UNSURE)},
  pages={74--83},
  year={2023},
  organization={Springer}
}

@InProceedings{tta2,
author="Wang, Guotai
and Li, Wenqi
and Ourselin, S{\'e}bastien
and Vercauteren, Tom",
editor="Crimi, Alessandro
and Bakas, Spyridon
and Kuijf, Hugo
and Keyvan, Farahani
and Reyes, Mauricio
and van Walsum, Theo",
title="Automatic Brain Tumor Segmentation Using Convolutional Neural Networks with Test-Time Augmentation",
booktitle="Brainlesion: Glioma, Multiple Sclerosis, Stroke and Traumatic Brain Injuries",
year="2019",
publisher="Springer International Publishing",
address="Cham",
pages="61--72",
}

\end{document}